\documentclass[11pt]{article}
\usepackage{authblk}
\usepackage[hyperref]{acl2020}
\usepackage{times}
\usepackage{url}
\usepackage{latexsym}
\usepackage{enumitem}

\usepackage{microtype}
\aclfinalcopy % Uncomment this line for the final submission
\usepackage{booktabs}
\usepackage{multirow}
\usepackage{adjustbox}
\usepackage{arydshln}
\usepackage{tabularx}
\usepackage{array}
\usepackage{graphicx,subcaption}
\usepackage{enumitem}
\usepackage{amsmath}

\title{Conversational Question Reformulation via Sequence-to-Sequence Architectures and Pretrained Language Models}

\author[1]{Sheng-Chieh Lin\thanks{\hspace{0.16cm}Contributed equally.}\hspace{0.14cm}}
\newcommand\CoAuthorMark{\footnotemark[\arabic{footnote}]} % get the current value
\author[1]{Jheng-Hong Yang\protect\CoAuthorMark \hspace{0.14cm}}
\author[2]{Rodrigo Nogueira}
\author[1]{\\Ming-Feng Tsai}
\author[1]{Chuan-Ju Wang}
\author[2]{Jimmy Lin}
\affil[1]{Research Center for Information Technology Innovation, Academia Sinica}
\affil[2]{David R. Cheriton School of Computer Science, University of Waterloo}

\date{}

\begin{document}
\maketitle
\begin{abstract}
  This paper presents an empirical study of conversational question reformulation (CQR) with sequence-to-sequence architectures and pretrained language models (PLMs).
  We leverage PLMs to address the strong token-to-token independence assumption made in the common objective, maximum likelihood estimation, for the CQR task.
  In CQR benchmarks of task-oriented dialogue systems, we evaluate fine-tuned PLMs on the recently-introduced CANARD dataset as an in-domain task and validate the models using data from the TREC 2019 CAsT Track as an out-domain task.
  Examining a variety of architectures with different numbers of parameters, we demonstrate that the recent text-to-text transfer transformer (T5) achieves the best results both on \mbox{CANARD} and CAsT with fewer parameters, compared to similar transformer architectures.
\end{abstract}

\section{Introduction}
\label{sec:intro}

Natural-language dialogue capabilities play an essential role as an enabling 
technology in intelligent personal assistants to understand and connect people~\cite{neural_conv_ai}.
Effective dialogue systems require many components, including natural language understanding, dialogue state tracking, and natural language generation~\cite{dialogue}.
Of late, practitioners in industry~\cite{cqu} and researchers in academia~\cite{canard} have made substantial progress in a variety of methods to improve end-to-end task-oriented dialogue systems.

\begin{figure}[t]
    \centering
    \resizebox{\columnwidth}{!}{
        \includegraphics{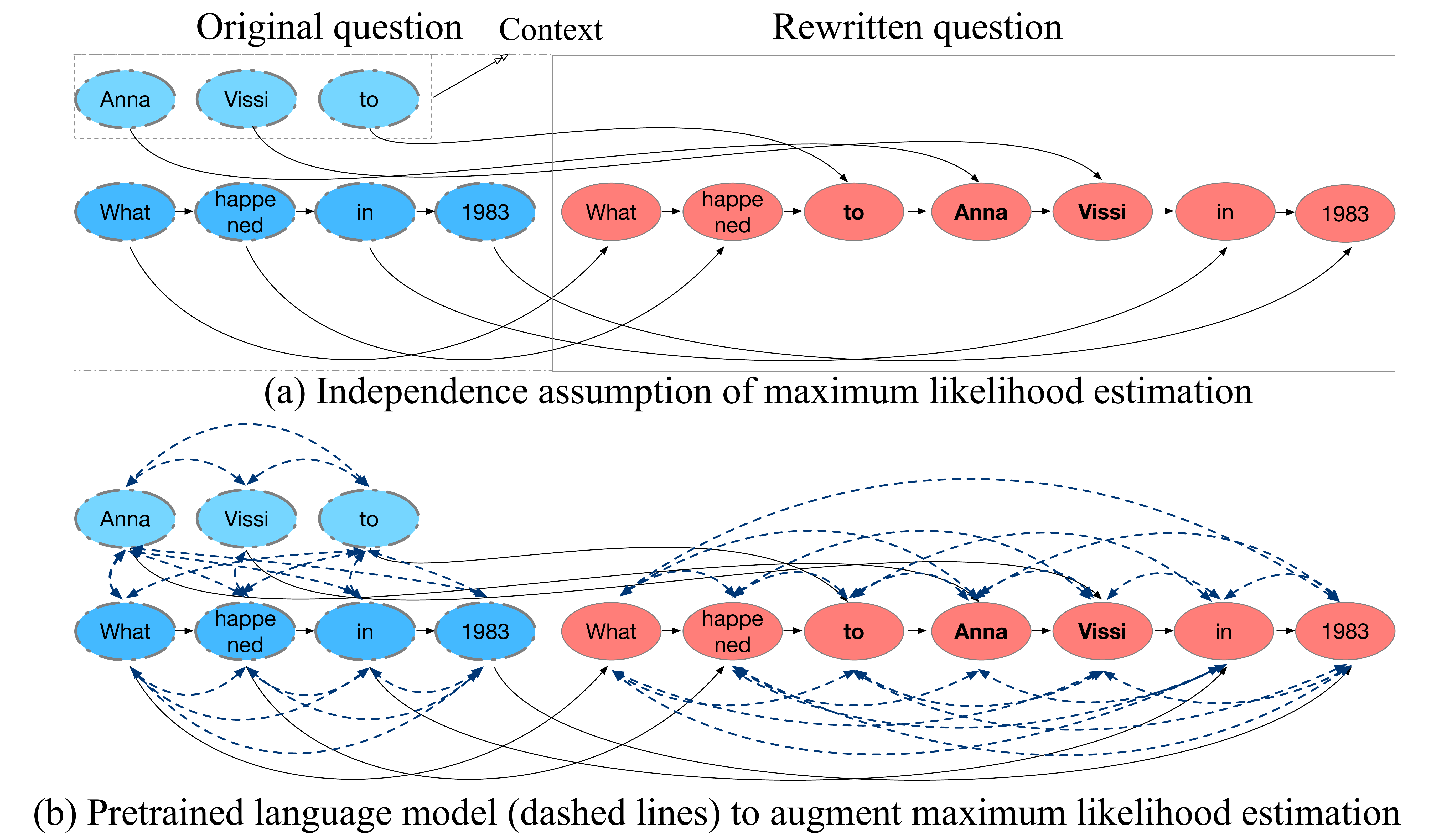}
    }
    \caption{Conversational question reformulation.}
    \vspace{-.2cm}
    \label{fig:idea}
\end{figure}

Due to the complex and nuanced nature of human communication, conversations often contain utterances that include coreference, ellipsis, and other phenomena; thus, a good dialogue system should be able to resolve these ambiguities to accurately reconstruct the user's original intent.
We present an example from~\citet{canard} in Figure~\ref{fig:idea} to illustrate the task of conversational question reformulation (CQR).

However, as we can observe from Figure~\ref{fig:idea}(a), applying maximum likelihood estimation (MLE) purely based on human-rewritten sentences introduces a strong independence assumption that does not consider conversation dependencies or linguistic structure.
Thanks to great progress made by language models pretrained on large corpora using self-supervised learning objectives~\cite{BERT, gpt2, unilm, t5}, there are now many models equipped with knowledge of various language structures extracted from human-generated texts.
We can leverage these models to relax the independence assumption in a pure MLE objective, shown in Figure~\ref{fig:idea}(b).

We list the contributions of this work as follows:

\begin{itemize}[leftmargin=*]
    \item We conduct, to our knowledge, the first empirical study leveraging pretrained language models to relax the independence assumption made in using an MLE objective in a CQR task.
    \item We achieve the state of the art in terms of BLEU on two CQR benchmarks of task-oriented dialogue systems:\ (a) conversational open-domain question answering with CANARD and (b) conversational search with TREC CAsT.
\end{itemize}

\noindent In summary, this work demonstrates a simple yet effective way to resolve coreference and ellipsis in a CQR task by leveraging pretrained language models.
Furthermore, among representative models, we find that a well-tuned text-to-text transfer transformer (T5) reaches performance that is on par with humans on the in-domain CANARD dataset and achieves the best performance on the out-of-domain CAsT dataset.

\section{Related Work}
\label{sec:related_work}
 
\textbf{Conversational search}~\cite{conv_search} covers a broad range of techniques that facilitate an IR task in a conversational context:\ natural language interactions, cumulative clarification~\cite{Aliannejadi2019AskingCQ}, feedback collection, and information needs profiling during conversations.
CQR is an important component of conversational search systems.
In order to resolve users' information needs to retrieve relevant answers, a CQR module that leverages pretrained models is a promising approach, compared to alternatives that track dialogue states based on ``cheap'' but noisy implicit feedback from users~\cite{uddin2018multitask, uddin2019context} or ``expensive'' but sparse judgments~\cite{cast}.

\smallskip \noindent
\textbf{Open-domain question answering (QA) systems} return answers in response to user questions, both in natural language, from a broad range of domains~\cite{sun-etal-2018-open}.
With great progress coming from contributions by the NLP and IR communities, high quality datasets for single-turn~\cite{squad, narrativeqA, quasar} and multi-turn (conversational)~\cite{coqa, quac} open-domain QA are available today.
These datasets have led to many successful supervised techniques for various tasks~\cite{drqa, bidaf, flowqa}.

Recently, to improve dialogue understanding, researchers have proposed collecting annotations on resolving multi-turn dialogues in the context of question answering tasks~\cite{cqu, canard}.
Building on this line of thought, our work addresses the problem of modeling question rewrites in multi-turn dialogues, especially in the context of open-domain conversational QA.

\section{Conversational Question Reformulation}
\label{sec:cqr}

\subsection{Problem Formulation}

We first formally define the conversational question reformulation (CQR) task, which is also called question de-contextualization~\cite{canard} or conversational question (query) understanding in the context of task-oriented dialogue systems~\cite{cqu}.
Consider a topic $t$ from a set of topics $T$, given a topic-oriented utterance sequence (i.e., the conversation history):\ $H^{t}= \{u_1,\cdots,u_{i},u_{i+1},\cdots u_{N} \}$ of $N$ utterances, each of which could be a question $q_i$ or an answer $a_i$ at the $i$-th turn.
The task is to reformulate the question $q_{i}$ into $\bar{q}_{i}$ that incorporates the context $H^{t}_{<i} = \{ u_{j} \}_{j=1}^{i-1}$.
In other words, we wish to automatically reformulate the input question $q_{i}$ by infusing information that exists in the context $H^t$ but is missing from the question itself.

Following the definitions of~\citet{cqu} and~\citet{canard}, we further refine the task scope of reformulating $\bar{q}_i$.
Given a question $q_i$ with its historical context $H^{t}_{<i}$ and a human-rewritten ground truth $\bar{q}_i$, our objective is to induce a function $ F(\{ q_i, H^{t}_{<i} \}) = \bar{q}_i$,
where $\bar{q}_i$ is comprised of tokens $\{ y_k \}_{k=1}^{m}$ of length $m$ from the context comprising the dialogue sequence $\{ q_i, H^{t}_{<i} \}$ (current and historical utterances), modeled as a sequence of tokens $\{ x_k \}_{k=1}^{n}$ of length $n$.
The tokens $y_k$'s can either be drawn from the context $H^{t}_{<i}$ or the current input $q_i$.
In the reconstruction of the ground truth, human annotators are asked to maintain the sentence structure of $q_i$ by copying phrases from the original utterances and performing as few edits as possible.

Finally, given probability $P$ conditioned on a parameterized function $\hat{F}$ and the context (current and historical utterances), the overall objective of the task is then defined in terms of finding the parameters $\theta$ by maximum likelihood estimation:
\begin{equation}
\small
\label{eq:objective}
    \theta = \mathop{\arg\max_{{\theta}}} \prod_{t=1}^{T} \prod_{i=1}^{N}
                P_{ti} \left( \bar{q}_i | \hat{F}( \{ q_i, H^{t}_{<i} \}, \theta )  \right).
\end{equation}

\subsection{Sequence-to-Sequence Architectures and Pretrained Language Models}

As both the input $q_i$ and the output $\bar{q}_i$ are posed in natural language, a reasonable choice for the parametric function is a sequence-to-sequence (S2S) model~\cite{seq2seq, transformer}.
With this design, we can incorporate context-dependent sentence-level structures when generating output tokens, since the model can consider both the previously-generated sequences as well as the context.

To extract information from the conversation flow, a simple approach, proposed by~\citet{xiong-etal-2018-session} and~\citet{canard}, is to concatenate the historical utterances $H^{t}_{<i}$ with the current input $q_i$ as the context $\left[ {H}^{t}_{<i}\,\Vert\, q_i \right]$, and then use a S2S model to infer the output sequence $\hat{q}_i$ based on it.
To optimize parameters in the S2S model, we can adopt a supervised learning approach to train the S2S model to generate the $\hat{q}_i$ tokens, given the $\bar{q}_i$ tokens as ground truth output.

However, Eq~(\ref{eq:objective}) makes an important assumption:\ here, we consider each conversation topic $t$ and each $i$-th turn independently.
Since a topic-oriented conversation is often coherent and smoothly spans several utterances, an approximation of the parameterized function $\hat{F}(\cdot, \theta)$ purely based on Eq~(\ref{eq:objective})\ could be sub-optimal.
To relax this assumption, we introduce pretrained language models~\cite{BERT, gpt2, t5} to leverage language structures extracted from large corpora.
Specifically, we adopt these models and fine-tune their pretrained weights, as in previous work~\cite{gpt2, t5}.

\section{Experiments}

\subsection{Dataset}

To evaluate the capability of various models in reformulating conversational questions, we conduct experiments on the CANARD dataset~\cite{canard}, an existing large open-domain dataset for CQR (containing over 30k training samples).
Each sample in the CANARD dataset includes an original query from the QuAC dataset~\cite{quac}, its context (historical utterances and their answers), and the corresponding rewritten question by human annotators.  

In addition, we also evaluate model performance on the dataset provided by the TREC~2019 Conversational Assistant Track (CAsT).\footnote{\url{https://github.com/daltonj/treccastweb}} CAsT organizers manually rewrote each conversational query in the evaluation set according to its contextual information and previous utterances in the same session. 
Statistics of the CANARD and CAsT datasets are presented in Table~\ref{tb:statisitics}. 

\begin{table}[t]
	\centering
	\small
	\caption{Statistics of the datasets used in this work.}
	\vspace{-.2cm}
    \begin{tabular}{cccc}
    \toprule
   	\multicolumn{3}{c}{CANARD}& \multirow{2}{*}{CAsT}\\
	\cmidrule(lr){1-3}
	    Train & Dev & Test  &  \\
	\midrule
	    31,538 & 3,418 & 5,571 & 479 \\
	\bottomrule
	\end{tabular}
	\vspace{-.2cm}
	\label{tb:statisitics}
\end{table}

\subsection{Setup}

To train and evaluate our sequence-to-sequence (S2S) models, we construct model input largely following \citet{canard}.
Specifically, we concatenate each original question and its context by adding special separator tokens between them.
Separator tokens are also added to contextual information to separate historical utterances.
The human-rewritten questions serve as the ground truth target sequences. 
For encoder- or decoder-only models (e.g., GPT-2, BERT, and UniLM), each training input sequence (as described above) is concatenated with its target sequence, and the models are trained to recover the target sequence using standard masking tricks.

We train each model on the CANARD training set and select the checkpoint with the best performance on development set.
In addition to comparing model performance on the CANARD test set, we directly use the model trained on CANARD to perform CQR on the CAsT dataset.\footnote{Note that for CAsT, only historical questions are included as contextual information.}
Model performance is computed by the BLEU score between model output and the human-rewritten ground truth. 
Table~\ref{tb:settings} shows the settings of the neural models. 

Additional model-specific training details are as follows.
(a)~\textbf{LSTM}:\
Following the script provided by~\citet{canard}, we train a bidirectional LSTM S2S model with attention; the word embeddings are initialized with GloVE.\footnote{\url{https://github.com/aagohary/canard}}
(b)~\textbf{GPT-2}~\cite{gpt2}, which can be characterized as a pretrained decoder-only transformer:\
To focus on rewriting questions, we fine-tune the model (GPT-2 medium) by masking the cross entropy loss at the positions of the contextual tokens.   
(c)~\textbf{BERT}~\cite{BERT}, which can be characterized as a pretrained encoder-only transformer:\
Following the S2S fine-tuning procedure proposed in~\citet{unilm}, we fine-tune BERT-large (cased) by randomly masking the tokens with 70\% probability in targeted sequences.\footnote{\url{https://github.com/microsoft/unilm}}
(d)~\textbf{UniLM}~\cite{unilm}, where the model architecture is the same as BERT large and pretrained using three types of language-modeling tasks:\
The method for fine-tuning is the same as BERT.
(e)~\textbf{T5}~\cite{t5}, an encoder--decoder transformer that maps natural language understanding tasks to text-to-text transformation tasks:\
We fine-tune the T5-base model with the same settings used in~\citet{doctttttquery}.\footnote{\url{https://github.com/castorini/docTTTTTquery}}

In addition, we list human performance of CQR (denoted as \textbf{Human}), as measured by \citet{canard}, and the baseline performance using questions without any reformulation (denoted as \textbf{Raw}) for comparison.

\begin{table}[t]
	\caption{Model settings. }
	\vspace{-.2cm}
	\label{tb:settings}
	\centering
	\small
	\resizebox{\columnwidth}{!}{
    \begin{tabular}{lccc}
	\toprule
	& \# parameters & Learning rate & Batch size\\
	\midrule
	   LSTM & 46M & $0.15$ & 16\\
	   GPT-2-medium & 345M & $10^{-4}$ & 32\\
       BERT-large & 340M &$10^{-5}$ & 32\\
       UniLM-large & 340M & $10^{-5}$ & 32 \\
       T5-base  & 220M &$10^{-4}$ & 256\\
	\bottomrule
	\end{tabular}
	}
	\vspace{-0.1cm}
\end{table}

\subsection{Results}

Our main results in terms of BLEU on CANARD and CAsT are shown in Table~\ref{tb:bleu}, using greedy search decoding for inference.
In general, all neural S2S models perform better than the original questions (Raw), except for LSTM on CAsT.
This indicates that the PLMs (GPT2, BERT, UniLM, and T5) have obtained at least some generalization capability on the CQR task.

Among all neural S2S models, T5 demonstrates a better ability to learn CQR from human-rewritten questions with fewer model parameters.
Specifically, in the CANARD test set, T5 beats the other neural S2S models with 58.08 BLEU, which is close to human performance, 59.92.
Furthermore, on CAsT, T5 achieves the highest BLEU score (75.07), four points better than the second-best model (71.21).
These results demonstrate T5's superior generalization ability. 

\begin{figure}[tb!]
\centering
    \includegraphics[width=\columnwidth]{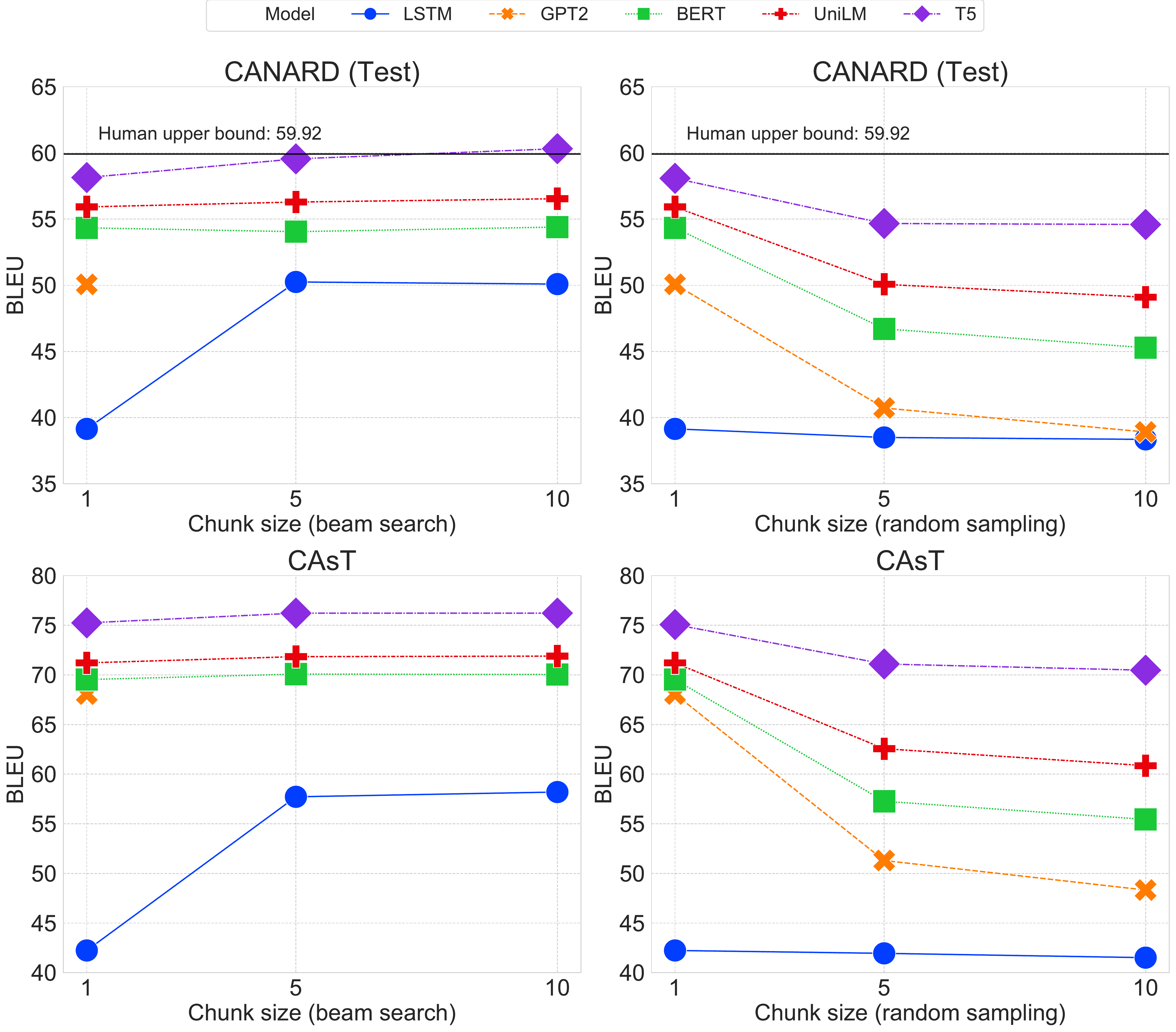}
\caption{Decoding sensitivity analysis}
\label{fig:sensitivity}
\vspace{-0.1cm}
\end{figure}

In addition, we also perform S2S model inference using beam search and top-$k$ random sampling decoding.\footnote{Note that beam search (top-$k$ random sampling) is equal to greedy search when the beam width (top-$k$) is set to 1}
Figure~\ref{fig:sensitivity} (left side) shows that beam search with larger beam widths further improves BLEU scores in both datasets.
T5 with a beam width of 10 achieves a BLEU score that is on par with human performance on the CANARD test set and reaches 76.22 on CAsT.\footnote{We did not perform GPT-2 inference with beam search since the original implementation does not support beam search.} 
Figure~\ref{fig:sensitivity} (right side) illustrates that random sampling with larger top-$k$ leads to poor BLEU scores.\footnote{For random sampling, we perform model inference with 10 repetitions and average over them.}
Under this decoding strategy, T5 still maintains better BLEU scores compared to the other S2S models.  
 
\begin{table}[t]
	\caption{BLEU score comparison. For simplicity, we compare neural S2S models using greedy search.}
	\label{tb:bleu}
	\centering
	\small
    \begin{tabular}{lccc}
	\toprule
	&\multicolumn{2}{c}{CANARD}& \multirow{2}{*}{CAsT}\\
	\cmidrule(lr){2-3}
	     & Dev  & Test  &  \\
	\midrule
	   Human &\multicolumn{2}{c}{59.92}&-\\
	   Raw & 33.84& 36.25& 60.41\\
       \midrule
	   LSTM & 43.68 & 39.15 & 42.24\\
	   GPT-2-medium &52.63 & 50.07& 68.07\\
       BERT-large &55.34 &54.34 &69.53\\
       UniLM-large & 57.39 & 55.92 & 71.21 \\
       T5-base  & \textbf{59.13} & \textbf{58.08} & \textbf{75.07}\\
	\bottomrule
	\end{tabular}
	\vspace{0.2cm}
\end{table}

\section{Conclusion}

In this paper, we conduct experiments on conversational question reformulation (CQR) via neural sequence-to-sequence (S2S) models and demonstrate that our fine-tuned T5-base model achieves the state of the art, in one case achieving performance on par with humans (at least measured by BLEU).
In addition, experiments on the CAsT dataset show that our fine-tuned T5-base model can be directly used in a transfer setting and beats other neural S2S models by quite a large margin.

\section{Acknowledgements}

This research was supported in part by the Canada First Research Excellence Fund and the Natural Sciences and Engineering Research Council (NSERC) of Canada.
Additionally, we would like to thank Google for computational resources in the form of Google Cloud credits.

\bibliographystyle{acl_natbib}
\bibliography{paper}

\begin{thebibliography}{26}
\expandafter\ifx\csname natexlab\endcsname\relax\def\natexlab#1{#1}\fi

\bibitem[{Ahmad et~al.(2018)Ahmad, Chang, and Wang}]{uddin2018multitask}
Wasi~Uddin Ahmad, Kai-Wei Chang, and Hongning Wang. 2018.
\newblock Multi-task learning for document ranking and query suggestion.
\newblock In \emph{Proc. ICLR}.

\bibitem[{Ahmad et~al.(2019)Ahmad, Chang, and Wang}]{uddin2019context}
Wasi~Uddin Ahmad, Kai-Wei Chang, and Hongning Wang. 2019.
\newblock Context attentive document ranking and query suggestion.
\newblock In \emph{Proc. SIGIR}, page 385–394.

\bibitem[{Aliannejadi et~al.(2019)Aliannejadi, Zamani, Crestani, and
  Croft}]{Aliannejadi2019AskingCQ}
Mohammad Aliannejadi, Hamed Zamani, Fabio Crestani, and W.~Bruce Croft. 2019.
\newblock Asking clarifying questions in open-domain information-seeking
  conversations.
\newblock In \emph{Proc. SIGIR}, page 475–484.

\bibitem[{Chen et~al.(2017)Chen, Fisch, Weston, and Bordes}]{drqa}
Danqi Chen, Adam Fisch, Jason Weston, and Antoine Bordes. 2017.
\newblock Reading {W}ikipedia to answer open-domain questions.
\newblock In \emph{Proc. ACL}, pages 1870--1879.

\bibitem[{Choi et~al.(2018)Choi, He, Iyyer, Yatskar, Yih, Choi, Liang, and
  Zettlemoyer}]{quac}
Eunsol Choi, He~He, Mohit Iyyer, Mark Yatskar, Wen-tau Yih, Yejin Choi, Percy
  Liang, and Luke Zettlemoyer. 2018.
\newblock {QuAC}: Question answering in context.
\newblock \emph{arXiv:1808.07036}.

\bibitem[{Devlin et~al.(2018)Devlin, Chang, Lee, and Toutanova}]{BERT}
Jacob Devlin, Ming-Wei Chang, Kenton Lee, and Kristina Toutanova. 2018.
\newblock {BERT}: Pre-training of deep bidirectional transformers for language
  understanding.
\newblock \emph{arXiv:1810.04805}.

\bibitem[{Dhingra et~al.(2017)Dhingra, Mazaitis, and Cohen}]{quasar}
Bhuwan Dhingra, Kathryn Mazaitis, and William~W. Cohen. 2017.
\newblock Quasar: Datasets for question answering by search and reading.
\newblock \emph{arXiv:1707.03904}.

\bibitem[{Dong et~al.(2019)Dong, Yang, Wang, Wei, Liu, Wang, Gao, Zhou, and
  Hon}]{unilm}
Li~Dong, Nan Yang, Wenhui Wang, Furu Wei, Xiaodong Liu, Yu~Wang, Jianfeng Gao,
  Ming Zhou, and Hsiao-Wuen Hon. 2019.
\newblock Unified language model pre-training for natural language
  understanding and generation.
\newblock In \emph{Proc. NIPS}.

\bibitem[{Elgohary et~al.(2019)Elgohary, Peskov, and Boyd-Graber}]{canard}
Ahmed Elgohary, Denis Peskov, and Jordan Boyd-Graber. 2019.
\newblock Can you unpack that?\ {Learning} to rewrite questions-in-context.
\newblock In \emph{Proc. EMNLP}, pages 5917--5923.

\bibitem[{Gao et~al.(2018)Gao, Galley, and Li}]{neural_conv_ai}
Jianfeng Gao, Michel Galley, and Lihong Li. 2018.
\newblock Neural approaches to conversational {AI}.
\newblock \emph{arXiv:1809.08267}.

\bibitem[{Huang et~al.(2019)Huang, Choi, and tau Yih}]{flowqa}
Hsin-Yuan Huang, Eunsol Choi, and Wen tau Yih. 2019.
\newblock Flow{QA}: Grasping flow in history for conversational machine
  comprehension.
\newblock In \emph{Proc. ICLR}.

\bibitem[{Jeffrey et~al.(2019)Jeffrey, Xiong, and Callan}]{cast}
Dalton Jeffrey, Chenyan Xiong, and Jamie Callan. 2019.
\newblock {CAsT 2019}: The conversational assistance track overview.
\newblock In \emph{Proc. TREC}.

\bibitem[{Ko{\v{c}}isk{\'y} et~al.(2018)Ko{\v{c}}isk{\'y}, Schwarz, Blunsom,
  Dyer, Hermann, Melis, and Grefenstette}]{narrativeqA}
Tom{\'a}{\v{s}} Ko{\v{c}}isk{\'y}, Jonathan Schwarz, Phil Blunsom, Chris Dyer,
  Karl~Moritz Hermann, G{\'a}bor Melis, and Edward Grefenstette. 2018.
\newblock The {N}arrative{QA} reading comprehension challenge.
\newblock \emph{Trans.of ACL}, 6:317--328.

\bibitem[{Nogueira and Lin(2019)}]{doctttttquery}
Rodrigo Nogueira and Jimmy Lin. 2019.
\newblock From doc2query to {docTTTTTquery}.

\bibitem[{Radford et~al.(2018)Radford, Wu, Child, Luan, Amodei, and
  Sutskever}]{gpt2}
Alec Radford, Jeffrey Wu, Rewon Child, David Luan, Dario Amodei, and Ilya
  Sutskever. 2018.
\newblock Language models are unsupervised multitask learners.

\bibitem[{Radlinski and Craswell(2017)}]{conv_search}
Filip Radlinski and Nick Craswell. 2017.
\newblock A theoretical framework for conversational search.
\newblock In \emph{Proc. CHIIR}, page 117–126.

\bibitem[{Raffel et~al.(2019)Raffel, Shazeer, Roberts, Lee, Narang, Matena,
  Zhou, Li, and Liu}]{t5}
Colin Raffel, Noam Shazeer, Adam Roberts, Katherine Lee, Sharan Narang, Michael
  Matena, Yanqi Zhou, Wei Li, and Peter~J. Liu. 2019.
\newblock Exploring the limits of transfer learning with a unified text-to-text
  transformer.
\newblock \emph{arXiv:1910.10683}.

\bibitem[{Rajpurkar et~al.(2018)Rajpurkar, Jia, and Liang}]{squad}
Pranav Rajpurkar, Robin Jia, and Percy Liang. 2018.
\newblock Know what you don't know:\ {Unanswerable} questions for {SQ}u{AD}.
\newblock In \emph{Proc. ACL}, pages 784--789.

\bibitem[{Reddy et~al.(2019)Reddy, Chen, and Manning}]{coqa}
Siva Reddy, Danqi Chen, and Christopher~D. Manning. 2019.
\newblock {CoQA}: A conversational question answering challenge.
\newblock \emph{Trans. of ACL}, 7:249--266.

\bibitem[{Ren et~al.(2018)Ren, Ni, Malik, and Ke}]{cqu}
Gary Ren, Xiaochuan Ni, Manish Malik, and Qifa Ke. 2018.
\newblock Conversational query understanding using sequence to sequence
  modeling.
\newblock In \emph{Proc. WWW}, page 1715–1724.

\bibitem[{Seo et~al.(2017)Seo, Kembhavi, Farhadi, and Hajishirzi}]{bidaf}
Min~Joon Seo, Aniruddha Kembhavi, Ali Farhadi, and Hannaneh Hajishirzi. 2017.
\newblock Bidirectional attention flow for machine comprehension.
\newblock In \emph{Proc. ICLR}.

\bibitem[{Sun et~al.(2018)Sun, Dhingra, Zaheer, Mazaitis, Salakhutdinov, and
  Cohen}]{sun-etal-2018-open}
Haitian Sun, Bhuwan Dhingra, Manzil Zaheer, Kathryn Mazaitis, Ruslan
  Salakhutdinov, and William Cohen. 2018.
\newblock Open domain question answering using early fusion of knowledge bases
  and text.
\newblock In \emph{Proc. EMNLP}, pages 4231--4242.

\bibitem[{Sutskever et~al.(2014)Sutskever, Vinyals, and Le}]{seq2seq}
Ilya Sutskever, Oriol Vinyals, and Quoc~V. Le. 2014.
\newblock Sequence to sequence learning with neural networks.
\newblock In \emph{Proc. NIPS}, page 3104–3112.

\bibitem[{Vaswani et~al.(2017)Vaswani, Shazeer, Parmar, Uszkoreit, Jones,
  Gomez, Kaiser, and Polosukhin}]{transformer}
Ashish Vaswani, Noam Shazeer, Niki Parmar, Jakob Uszkoreit, Llion Jones,
  Aidan~N. Gomez, {\L{}ukasz}~Kaiser, and Illia Polosukhin. 2017.
\newblock Attention is all you need.
\newblock In \emph{Proc. NIPS}, pages 5998--6008.

\bibitem[{Xiong et~al.(2018)Xiong, Wu, Zhang, and
  Stolcke}]{xiong-etal-2018-session}
Wayne Xiong, Lingfeng Wu, Jun Zhang, and Andreas Stolcke. 2018.
\newblock Session-level language modeling for conversational speech.
\newblock In \emph{Proc. EMNLP}, pages 2764--2768.

\bibitem[{Zhao and Eskenazi(2016)}]{dialogue}
Tiancheng Zhao and Maxine Eskenazi. 2016.
\newblock Towards end-to-end learning for dialog state tracking and management
  using deep reinforcement learning.
\newblock In \emph{Proc. of SIGDIAL}, pages 1--10.

\end{thebibliography}

\end{document}